\documentclass[letterpaper, 10 pt, conference]{ieeeconf}  

\IEEEoverridecommandlockouts                              

\usepackage{graphics} 
\usepackage{epsfig} 
\usepackage{mathptmx} 
\usepackage{times} 
\usepackage{amsmath} 
\usepackage{amssymb}  

\usepackage{graphicx}
\usepackage{algorithm} 
\usepackage{algpseudocode} 
\usepackage{makecell}
\usepackage{booktabs}
\usepackage{cuted} 
\usepackage{caption} 
\newcommand{\argmin}{\operatornamewithlimits{argmin}}

\usepackage{xcolor}
\usepackage{multicol}
\definecolor{mycitecolor}{RGB}{71, 191, 38}
\definecolor{mylinkcolor}{RGB}{40, 115, 201}
\definecolor{DGreen}{RGB}{107, 190, 35}
\definecolor{darkblue}{rgb}{0.1, 0.1, 1}

\makeatletter
\let\NAT@parse\undefined
\makeatother
\usepackage[colorlinks=true, citecolor=mycitecolor,linkcolor=mylinkcolor,urlcolor=mycitecolor]{hyperref}

\title{\LARGE \bf
Distributed NeRF Learning for Collaborative Multi-Robot Perception
}

\author{Hongrui Zhao$^{1}$, Boris Ivanovic$^{2}$, Negar Mehr$^{3}$ 
\thanks{*This work is supported by the National Science Foundation, under grants ECCS-2145134 CAREER Award, CNS-2423130, and CCF-2423131}
\thanks{$^{1}$Hongrui Zhao is with Department of Aerospace Engineering, University of Illinois Urbana-Champaign
        {\tt\small hongrui5@illinois.edu}}%
\thanks{$^{2}$Boris Ivanovic is with NVIDIA 
        {\tt\small bivanovic@nvidia.com}}%
\thanks{$^{3}$Negar Mehr with Department of Mechanical Engineering, University of California, Berkeley
        {\tt\small negar@berkeley.edu}}%
}

\begin{document}

\maketitle
\thispagestyle{empty}
\pagestyle{empty}

\begin{strip}
\begin{minipage}{\textwidth}\centering
\vspace{-70pt}
\includegraphics[trim={0cm 7.2cm 0cm 0cm},clip,width=\textwidth]{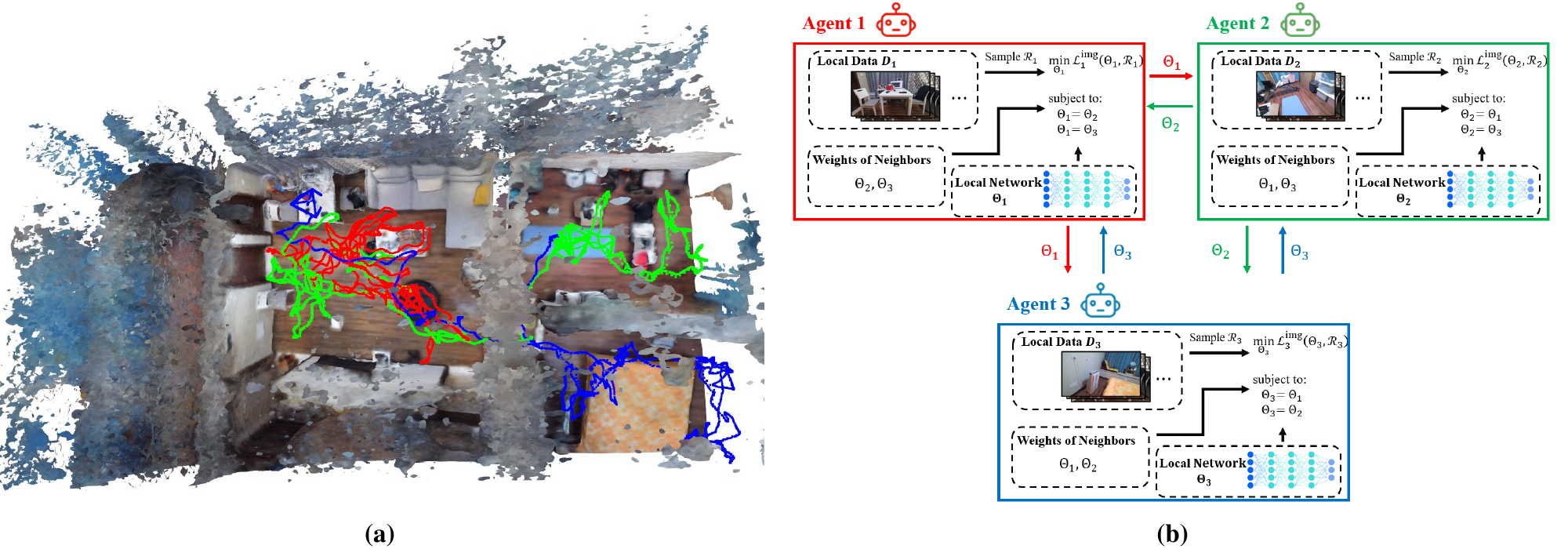}
\captionof{figure}{  (a) Three robotic agents explore a large-scale scene, with each agent's trajectory shown in a different color. 
The mesh, generated from the NeRF model of agent 1 (red), is displayed. 
Our distributed learning drives the model of agent 1 to reach consensus with the models of agents 2 and 3. 
The model represents the entire scene, and a complete mesh can be generated from it including regions agent 1 did not visit (the two rooms on the right). 
(b) In our distributed learning approach, only network weights $\Theta_i$ are shared among the agents.
Each agent minimizes the NeRF image reconstruction loss $L^{\text{img}}_i$ with mini-batch $R_i$ sampled from its local data $D_i$ while maintaining consensus with other agents.
This enables the distributed learning of a comprehensive scene representation without transferring raw data, making it suitable for multi-agent systems with limited communication bandwidth.}
\label{fig:large_traj}
\end{minipage}
\end{strip}


\begin{abstract}
Effective environment perception is crucial for enabling downstream robotic applications.
Individual robotic agents often face occlusion and limited visibility issues, whereas multi-agent systems can offer a more comprehensive mapping of the environment, quicker coverage, and increased fault tolerance. 
In this paper, we propose a collaborative multi-agent perception system where agents collectively learn a neural radiance field (NeRF) from posed RGB images to represent a scene. 
Each agent processes its local sensory data and shares only its learned NeRF model with other agents, reducing communication overhead.
Given NeRF's low memory footprint, this approach is well-suited for robotic systems with limited bandwidth, where transmitting all raw data is impractical.
Our distributed learning framework ensures consistency across agents' local NeRF models, enabling convergence to a unified scene representation.
We show the effectiveness of our method through an extensive set of experiments on datasets containing challenging real-world scenes, achieving performance comparable to centralized mapping of the environment where data is sent to a central server for processing.
Additionally, we find that multi-agent learning provides regularization benefits, improving geometric consistency in scenarios with sparse input views. We show that in such scenarios, multi-agent mapping can even outperform centralized training.
\end{abstract}

\section{INTRODUCTION}

Obtaining an accurate and comprehensive representation of the environment is a critical first step to enabling autonomous mobile robots to navigate and operate effectively.
Perceiving large-scale and complex environments using a single robotic agent is often inefficient due to limited sensor range and fields of view, making the process time-consuming. 
Relying on a single agent also creates a single point of failure, making the system vulnerable.
To address these challenges, multi-agent collaborative perception has emerged as a more efficient and robust solution. 
Multi-robot perception has demonstrated promising performance across various applications, including autonomous driving \cite{drive1,drive2,drive3}, warehouse automation \cite{warehouse}, indoor and outdoor scene reconstruction \cite{indoor1,indoor2}, and multi-UAV search and survey missions \cite{UAV}. 

A multi-agent perception system requires an efficient mechanism to aggregate sensory inputs from multiple robots into a unified, coherent representation. 
Recently, Neural Radiance Field (NeRF) \cite{NeRF} and its variants have gained significant attention in robotics research, and NeRF's density field has been successfully applied in robot motion planning and trajectory optimization \cite{NeRFrobot1,NeRFrobot2, NeRFRobot3, NeRFRobot4}. 
While NeRF applications for single robotic agents are well-researched, collaborative multi-agent NeRF reconstruction remains largely unexplored.
NeRF takes posed RGB images as input to train a multi-layer perceptron (MLP) that predicts the volume density and color of 3D points.
The neural network implicitly represents complex scene geometry and appearance with a low memory footprint, making NeRF well-suited for multi-agent systems with limited onboard storage and communication bandwidth. 
However, since each agent only has access to locally-captured sensory data, the primary challenge is efficiently merging local NeRF models while minimizing data transfer.

A straightforward approach to collaborative perception is to transfer raw images from all agents to a central server, where a single NeRF model is trained. 
However, this method may be impractical in real-world systems with limited communication bandwidth and is susceptible to communication disruptions and central server failures. 
Inspired by recent advances in distributed optimization \cite{DiNNO}, we propose an alternative approach in which each agent only communicates its neural network weights to other agents. 
Each agent uses its local data along with the network weights from its peers. By minimizing image reconstruction loss and maintaining consensus with other agents, each agent ultimately learns a NeRF model that captures the entire scene. 
Our experimental results demonstrate that this approach notably reduces data transfer compared to centralized training.

Our study also shows that distributing data across multiple agents can serve as an effective regularization technique, reducing NeRF overfitting compared to centralized training, where a single agent has access to the entire dataset. 
A challenge with NeRF is its requirement for dense scene coverage to achieve high-quality reconstruction. 
In real-world robotic applications, agents may only be able to observe the scene from limited viewpoints due to mobility or time constraints. 
With sparse inputs, NeRF tends to overfit to training images, resulting in inaccurate scene geometry and poor rendering quality from novel views~\cite{depthNeRF, RegNeRF}. 
This overfitting can also produce artifacts known as ``floaters'' near the camera, significantly degrading scene representation quality \cite{FreeNeRF}. 
In our approach, the requirement for consensus serves as a form of regularization, encouraging neural networks to achieve better generalization performance.

Our key contributions are as follows: 
\begin{enumerate}
\def\labelenumi{\arabic{enumi}.}
\item We propose a multi-agent collaborative learning approach for robotic perception, wherein all agents jointly learn a NeRF of the scene without transferring raw data.
\item We show that distributing data across multiple agents and enforcing consensus on the NeRF network weights reduces overfitting in sparse-view scenarios, compared to centralized training. 
\item We evaluate our multi-agent approach on several datasets, demonstrating that it performs comparably to centralized training and even surpasses it when input views are limited.
\item We demonstrate that our approach is more communication-efficient than centralized training and investigate its performance under varying communication disruptions.
\end{enumerate}


\section{RELATED WORKS}\label{Related Works}
\textbf{Multi-agent Perception}. 
Most existing multi-agent perception methods rely on transmitting either explicit local maps or raw sensor data to a central server. 
Typically, each agent maintains a set of 3D points with features extracted from images and sends this data to a central server for map alignment and fusion \cite{indoor1,CP-SLAM, activeMapping}. Alternatively, approaches such as MAANS \cite{MAANS} train each agent to predict bird's-eye view images of the environment, which are then centrally-fused for a more comprehensive scene representation. 
Another method, CVIDS \cite{indoor2}, involves agents directly transmitting raw data to a central server, which then aligns local coordinate systems and estimates a global truncated signed distance function (TSDF) map. 
However, these methods may be impractical for multi-agent systems with limited communication bandwidth. 
Despite efforts to reduce the amount of sensory data communicated~\cite{com}, leveraging neural implicit representations remains largely unexplored. This paper aims to fill this gap.

\textbf{Single-agent NeRF-based Perception}.
NeRF-based perception methods utilize RGB or RGBD images to train MLP networks for scene prediction tasks, such as color, volume density, occupancy, or signed distance. 
For example, iSDF \cite{isdf} employs RGBD images for real-time signed distance field reconstruction, while H2-Mapping \cite{H2-Mapping} uses two MLP networks for color and signed distance prediction, enabling real-time reconstruction on edge devices. 
GO-Surf \cite{wang2022go-surf} employs multi-resolution feature grids and shallow MLP networks for fast surface reconstruction. 
Works like iMAP \cite{iMAP} and NICE-SLAM \cite{NICE-SLAM} integrate camera pose tracking with scene representation, while Co-SLAM \cite{wang2023coslam} further improves reconstruction speed. 
These methods primarily focus on centralized training with a single agent using all data.
Building on prior research, our approach introduces a distributed training framework that can be effectively applied to multi-robot systems.

\textbf{Single-agent NeRF with Sparse Input Views}.
In real-world applications like AR/VR, autonomous driving, and robotics, NeRF typically only has access to a limited number of views of the target scene. 
In this sparse setting, NeRF often faces challenges with overfitting to observed images, leading to inaccuracies in the reconstructed scene geometry. 
Prior works tackling this issue can be broadly categorized into three main approaches: 
(1)~Incorporating depth information to assist in the reconstruction process \cite{depthNeRF, sparsenerf};
(2)~Leveraging pre-trained models to offer regularization or to extract features for input to NeRF \cite{RegNeRF, DietNeRF, yu2020pixelnerf, mvsnerf}; and 
(3)~Applying explicit geometry regularization guided by human intuition, such as enforcing smoothness in neighboring depth estimations and sparsity in density predictions \cite{RegNeRF, kim2022infonerf, DiffNeRF}.
In comparison to these methods, our multi-agent approach stands out in its ability to reduce overfitting without the need for additional data or pre-trained models.

\textbf{Distributed Neural Network Optimization}.
Distributed neural network optimization enables multiple agents to collaboratively optimize a deep neural network in a decentralized manner \cite{distributed}. 
Distributed stochastic gradient descent (DSGD) \cite{DSGD} updates local parameters by combining neighbors' parameters with local gradients, while distributed stochastic gradient tracking (DSGT) \cite{DSGT} enhances convergence by estimating the global gradient. 
Decentralized linearized alternating direction method (DLM) \cite{DLM} minimizes local cost functions using linearized objectives, and consensus alternating direction method of multipliers (C-ADMM) \cite{DiNNO} optimizes a neural network by approximating an augmented Lagrangian. 
In this paper, we utilize C-ADMM due to its superior performance in multi-robot tasks.

\section{BACKGROUND: SINGLE-AGENT NERF}\label{NeRF}
NeRF \cite{NeRF} uses an MLP network $F_{\Theta}$ as an implicit representation to store geometry and appearance information of a scene.
A set of pixels are sampled from captured RGB images to train the network $F_{\Theta}$.
With the camera intrinsic calibration matrix $K$ and pose $P$, each sampled pixel with 2D image plane coordinates $(u,v)$ is associated with a ray $\mathbf{r}$ in the direction $\mathbf{d}=PK^{-1}(u,v,1)^\top$.
A 3D location $\mathbf{r}(t) = \mathbf{o} + t\mathbf{d}$ is sampled along the ray $\mathbf{r}$, where $\mathbf{o}$ is the camera origin acquired from $K$, and $t\in [t_n, t_f]$ is the distance from the origin with $t_n$ and $t_f$ as near and far bounds of the scene.
Before being passed to $F_{\Theta}$, a positional encoding $\gamma(\cdot)$ is applied to both $\mathbf{r}(t)$ and $\mathbf{d}$:
\begin{equation}
	\gamma(\mathbf{x}) = \left(  \sin(2^0\mathbf{x}) , \cos(2^0\mathbf{x}), \cdots, \sin(2^{E-1}\mathbf{x}) , \cos(2^{E-1}\mathbf{x})  \right),
\end{equation}
where $x$ could either be a 3D location $\mathbf{r}(t)$ or directional vector $\mathbf{d}$ and $E$ controls the encoding frequencies. 
A larger $E$ encourages the model to capture higher-frequency components in the input images, and $E$ may differ for $\mathbf{r}(t)$ and $\mathbf{d}$.
The positional encoding $\gamma(\cdot)$ is a mapping from $\mathbb{R}^3$ to $\mathbb{R}^{2E}$, allowing NeRF to better represent scenes with high-frequency variations in color and geometry. 
The MLP maps the encoded location and direction to a color $\mathbf{c}$ and a volume density $\sigma$:
\begin{equation}
	F_{\Theta} : \left( \gamma(\mathbf{r}(t)), \gamma(\mathbf{d}) \right) \to (\mathbf{c}, \sigma)
\end{equation}

With the color and volume density predictions from $F_{\Theta}$, the estimated color $\hat{\mathbf{C}}(\mathbf{r})$ of ray $\mathbf{r}$ is: 
\begin{equation}
	\hat{\mathbf{C}}(\mathbf{r}) = \int_{t_n}^{t_f}   T(t) \sigma(\mathbf{r}(t)) \mathbf{c}(\mathbf{r}(t),\mathbf{d})  \,dt 
\end{equation}
where $T(t)= \exp\left( -\int_{t_n}^{t} \sigma(\mathbf{r}(t))   \,ds   \right) $ denotes the accumulated transmittance along the ray from $t_n$ to $t$. 
If the transmittance $T(t)$ is low, then density and color rendered at $t$ will exert minimal impact on the integrated color $\hat{\mathbf{C}}(\mathbf{r})$.
The following image reconstruction loss is used for training the NeRF:
\begin{equation}
	L^{\text{img}}  = \sum_{\mathbf{r} \in R} \| \hat{\mathbf{C}}(\mathbf{r}) - \mathbf{C}(\mathbf{r})  \|_2^2 
\end{equation}
where $R$ represents a mini-batch of rays, and $\mathbf{C}(\mathbf{r})$ is the ground truth RGB color for ray $\mathbf{r}$.

\section{MULTI-AGENT PERCEPTION WITH NERF} \label{Multi-Agent Perception with NeRF}

An illustration of our multi-agent system is shown in Fig. \ref{fig:large_traj}.
The agents operate within a communication graph $G=(\mathcal{V}, \mathcal{E})$. 
Each agent is represented as a node $i \in \mathcal{V}$, and their connectivity is defined by a set of edges $\mathcal{E}$. 
Agents share their network weights $\Theta_i$ with their communication neighbors $N_i = \{ j \mid (i,j) \in \mathcal{E} \}$.
Each agent $i$ has access to its local data $D_i$, comprising images captured by the agent's onboard camera. 
During each optimization iteration, agent $i$ samples a batch of rays $R_i$ from $D_i$, and minimizes its local NeRF image reconstruction loss $L^{\text{img}}_i$. 
Additionally, each agent $i$ maintains consensus with its neighbors by aligning its NeRF model weights $\Theta_i$ with those of its neighbors $\Theta_j$ for all $j \in N_i$. 
This optimization process is formalized as:
\begin{subequations}
    \begin{align}
        \min_{\Theta_1, \Theta_2, \cdots} &\; \sum_{i \in \mathcal{V}}  L^{\text{img}}_i(\Theta_i, R_i)   \\
        \text{subject to} &\; \Theta_i = \Theta_j \; \forall (i,j) \in \mathcal{E} 
    \end{align}
    \label{opt_prob}
\end{subequations}
By minimizing their local loss functions and maintaining consensus with their neighbors, each agent's local model converges to a shared, global solution. 
This allows each agent to learn a comprehensive representation of the entire scene (the global solution) without transferring raw data, significantly improving communication efficiency.

Inspired by existing works such as DiNNO \cite{DiNNO}, we solve \eqref{opt_prob} in a distributed fashion using the C-ADMM algorithm, known for its superior convergence performance. 
For each agent $i$, we let $\Theta_i^k$, which is the network weights of agent $i$ at iteration $k$, be our primal variable and let $\mathbf{p}_i^k$ be the dual variable of agent $i$. 
The dual variables enforce consensus between agent $i$ and its communication neighbors. 
The entire process is detailed in Algorithm \ref{procedure}.
After initialization, each agent $i$ starts its main optimization loop. 
At each optimization iteration $k$ of the total $K$ iterations, each agent $i$ first exchanges its primal variable $\Theta_i^k$ with its neighbors $N_i$.
Then, both primal and dual variables are updated as follows:
\begin{subequations}
	\begin{align}
		\mathbf{p}_i^{k+1} &= \mathbf{p}_i^k + \rho \sum_{j \in N_i} ( \Theta_i^k - \Theta_j^k)  \label{dual_update}\\
		\Theta_i^{k+1} &= \argmin_{\Theta} \; L^{\text{img}}_i(\Theta, R_i) + \Theta^\top\mathbf{p}_i^{k+1} \nonumber \\
		&\hspace{1.5cm} + \rho \sum_{j \in N_i} \left \Vert  \Theta - \frac{\Theta_i^k + \Theta_j^k}{2} \right \Vert_2^2 , \label{primal_update}
	\end{align}
\end{subequations}
where $\rho$ is a fixed positive scalar controlling the step size  of the dual variable update. 
Note that, instead of the exact minimization in \eqref{primal_update}, we update the primal variable by taking $B$ steps of stochastic gradient descent to obtain $\Theta_i^{k+1}$, where $B$ is a user-defined constant. 

\begin{algorithm} 
	\caption{Multi-agent Distributed NeRF Learning}
	\label{procedure}
	\begin{algorithmic}[1]
		
		\For{$i \in \mathcal{V}$}
			\State $\mathbf{p}_i^{0} = 0$
			\State $\Theta_i^0 = \Theta_{\text{initial}} $ 
		\EndFor 

		\For{$k \leftarrow 0 \; \text{to} \; K$}
			\For{$i \in \mathcal{V}$}
				\State Send $\Theta_i^k$ to neighbors $N_i$
			\EndFor 

			\For{$i \in \mathcal{V}$}
				\State Perform dual variable update\eqref{dual_update}
				\State Obtain $\mathbf{p}_i^{k+1}$
				\For{$b \leftarrow 0 \; \text{to} \; B$}
					\State Sample $R_i$ from $D_i$
					\State Perform gradient descent for primal variable update\eqref{primal_update}
				\EndFor
				\State Obtain $\Theta_i^{k+1}$
			\EndFor 

			\EndFor
	\end{algorithmic}
\end{algorithm}

\section{EXPERIMENTAL RESULTS} \label{Experimental Results}
\textbf{Datasets.} We evaluate our algorithm on a number of datadsets.
The VR-NeRF Eyeful Tower \cite{VRNeRF} and NICE-SLAM Apartment \cite{NICE-SLAM} datasets, which feature large-scale real-world indoor scenes, are used to assess performance in complex environments. 
To investigate the impact of communication disruptions, we use the Replica dataset \cite{replica19arxiv}, whose smaller simulated scenes allow for comparable reconstruction results even with disrupted communications. 
For evaluating NeRF performance with sparse input views, we utilize the Blender \cite{NeRF} and LLFF \cite{LLFF} datasets. 
These two datasets, providing multi-view images of synthetic and real-world objects, enable learning geometry from limited views with effective regularization.

\textbf{Metrics.}
In all experiments, we compare the scene reconstruction performance of a single agent with full data access in a centralized setup against our multi-agent method, where each agent only accesses a subset of the data. 
For the Blender \cite{NeRF} and LLFF \cite{LLFF} datasets, we evaluate our method using the PSNR, SSIM \cite{SSIM}, and LPIPS \cite{LPIPS} metrics. 
PSNR measures reconstruction quality, with higher values indicating better performance. 
SSIM evaluates structural similarity, ranging from 0 to 1, where 1 indicates perfect similarity. 
LPIPS captures perceptual differences, with lower values indicating greater similarity. 


\subsection{VR-NeRF Eyeful Tower Dataset Results}
We first evaluate our method's ability to capture the detailed geometry of complex scenes using the Eyeful Tower dataset. 
We implement our distributed learning approach with nerf-pytorch \cite{lin2020nerfpytorch}, trained on a set of 12 images captured from various locations. 
In the centralized setup, NeRF has access to all 12 images. In the multi-agent setup, images were evenly distributed among agents. 
We assess the trained models using 12 novel-view test images and report performance for various numbers of agents.

\begin{figure}[hbt!]
	\centerline{\includegraphics[trim={5cm 1cm 3cm 0.5cm},clip,width=\columnwidth]{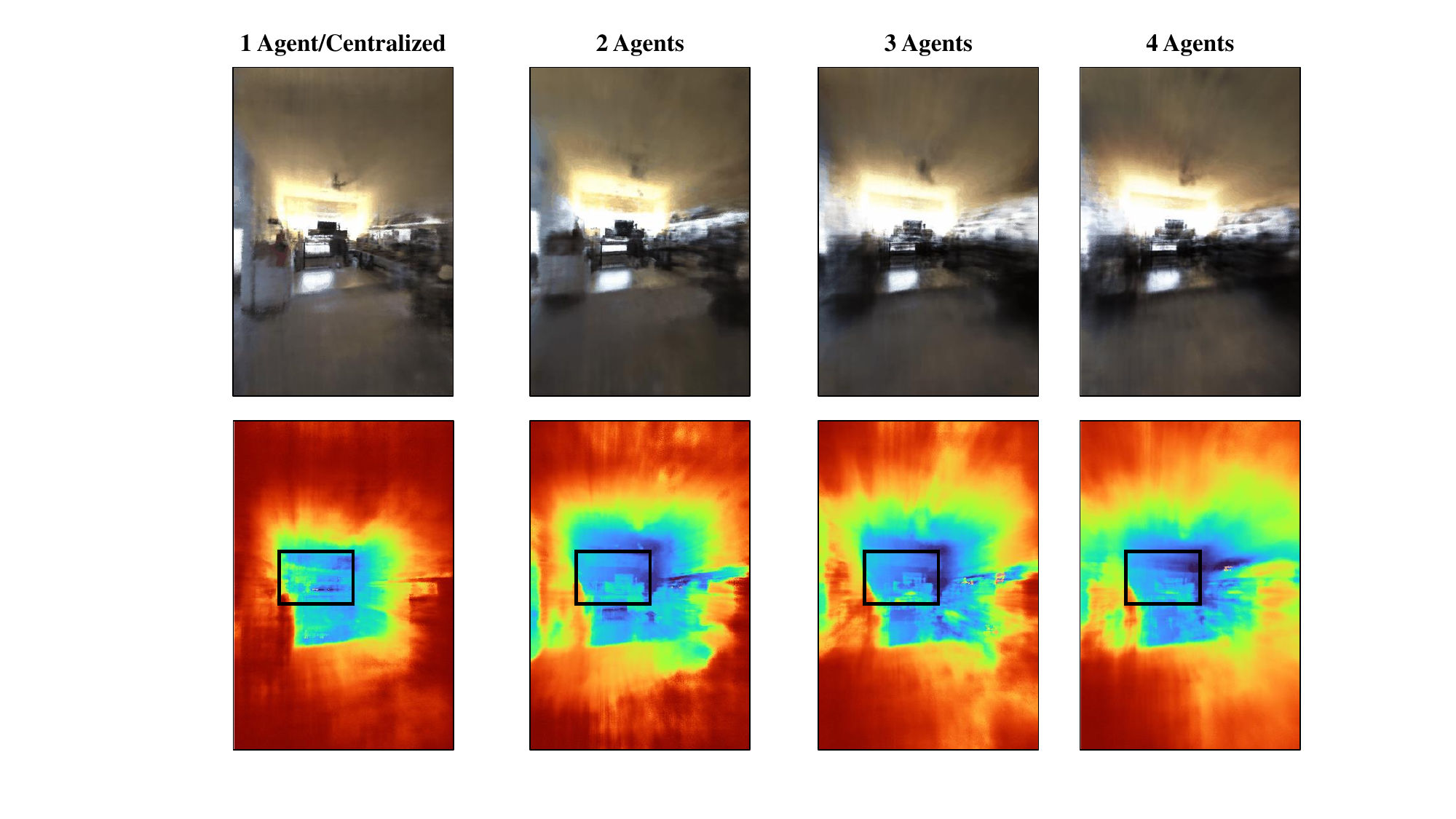}} 
	\caption{Our multi-agent approach reconstructs the complex geometry of indoor scenes with comparable quality to the centralized method. 
	In certain instances, the multi-agent approach captures high-frequency details more effectively than the centralized method, highlighted with black rectangles.}
	\label{fig3:VRNeRF}
\end{figure}

Our quantitative evaluations are summarized in Table \ref{tab:VRNeRF}, with selected rendered test images in Fig.~\ref{fig3:VRNeRF}.
Centralized training, which uses all raw data, provides an ideal upper bound for comparison. 
The multi-agent method achieved results comparable to centralized training across all three evaluation metrics. 
Notably, performance remained stable with varying numbers of agents. 
In some cases, the multi-agent approach captured high-frequency details better than the centralized method, highlighted with black rectangles in Fig. \ref{fig3:VRNeRF}. 
These results demonstrate that multiple agents can collaboratively learn detailed representations of complex indoor scenes, achieving fidelity comparable to centralized training \emph{without} sharing local images.

\begin{table}[hbt!]
  \caption{Eyeful Tower Dataset Results}
  \centering
  \resizebox{\columnwidth}{!}{%
    \begin{tabular}{c ccc ccc ccc}
      \toprule
      & \multicolumn{3}{c}{Apartment} & \multicolumn{3}{c}{Workshop} & \multicolumn{3}{c}{Furnished Room} \\
      \cmidrule(lr){2-4} \cmidrule(lr){5-7} \cmidrule(lr){8-10}
      \# Agents & PSNR $\uparrow$ & SSIM $\uparrow$ & LPIPS $\downarrow$ & PSNR $\uparrow$ & SSIM $\uparrow$ & LPIPS $\downarrow$ & PSNR $\uparrow$ & SSIM $\uparrow$ & LPIPS $\downarrow$ \\
      \midrule
      1 & \color{DGreen}\textbf{17.25} & \color{DGreen}\textbf{0.631} & \color{DGreen}\textbf{0.472} & \color{DGreen}\textbf{16.67} & \color{DGreen}\textbf{0.460} & \color{DGreen}\textbf{0.594} & 15.34 & 0.607 & \color{DGreen}\textbf{0.656} \\
      2 & 16.93 & 0.624 & 0.522 & 16.51 & 0.455 & 0.663 & \color{DGreen}\textbf{15.39} & \color{DGreen}\textbf{0.607} & 0.693 \\
      3 & 15.95 & 0.620 & 0.544 & 15.23 & 0.431 & 0.719 & 14.29 & 0.551 & 0.760 \\
      4 & 15.79 & 0.615 & 0.553 & 15.05 & 0.436 & 0.712 & 14.30 & 0.584 & 0.745 \\
      \bottomrule
    \end{tabular}%
  }
  \label{tab:VRNeRF}
\end{table}

\subsection{NICE-SLAM Apartment Dataset Results}
To assess large-scale reconstruction capabilities, we conduct experiments in a spacious indoor apartment (14.5m$\times$7.5m$\times$3.8m), where three agents explore different areas and capture extensive image data along distinct trajectories, illustrated in Figure \ref{fig:large_traj}. 
In total, the agents collected 12,594 pairs of RGB and depth images. Given vanilla NeRF's limited scalability, we apply our multi-agent training method to Co-SLAM \cite{wang2023coslam}, an advanced NeRF-based mapping method designed for larger scenes.

\begin{figure}[hbt!]
	\centerline{\includegraphics[trim={0cm 1cm 3cm 0cm},clip,width=\columnwidth]{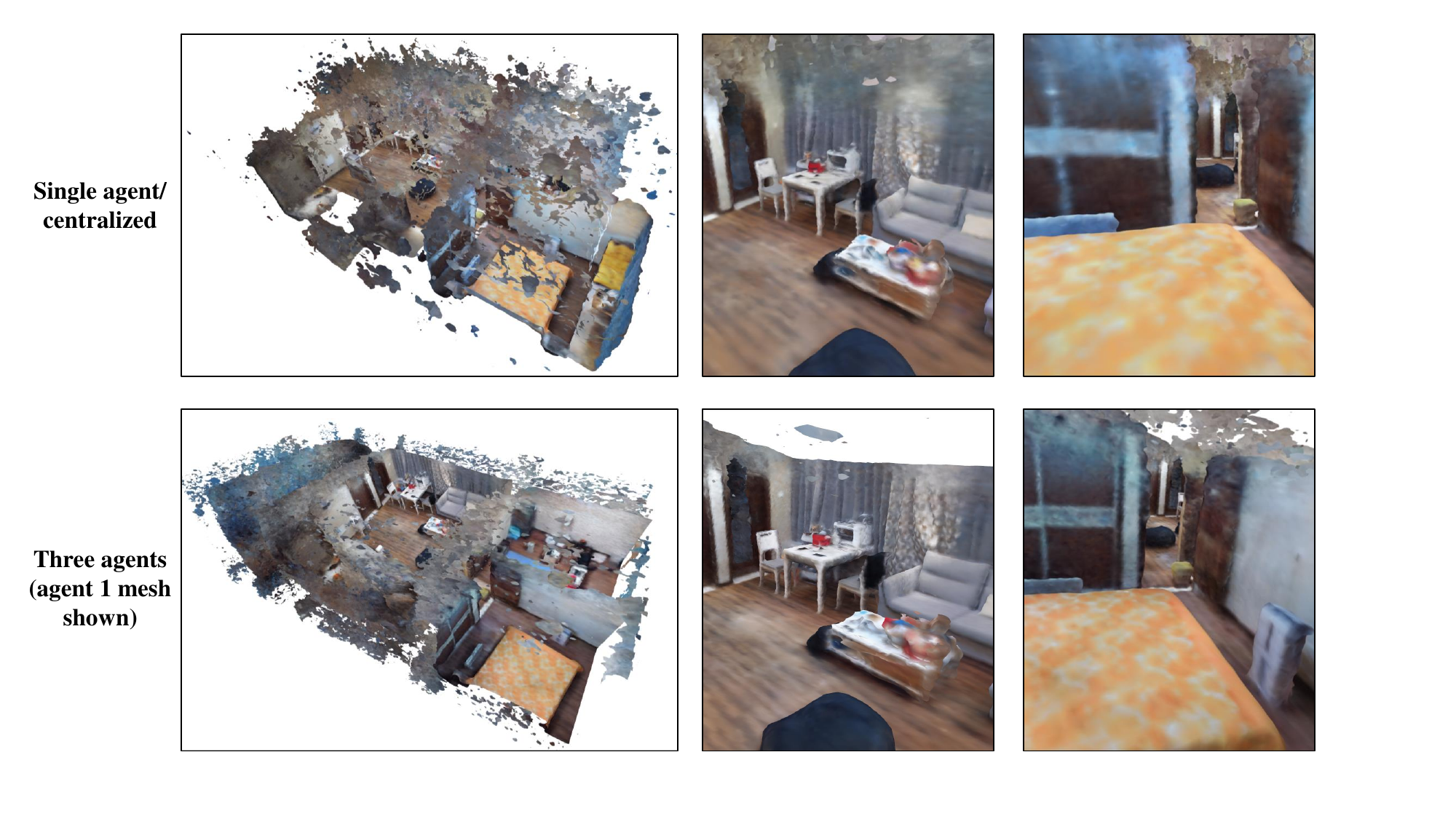}} 
	\caption{Comparison of scene reconstruction quality between centralized training and three-agent training. The multi-agent approach achieves performance comparable to the centralized method, with fewer rendering artifacts (``floaters"). These results highlight that our multi-agent learning method can handle larger scenes effectively and achieve a high level of reconstruction completeness}
	\label{fig:large_mesh}
\end{figure}

The experimental results are shown in Figure \ref{fig:large_mesh}. 
Rather than rendering individual images, we generate a 3D mesh of the entire scene to evaluate reconstruction completeness.
Overall, our multi-agent training achieves comparable reconstruction quality to the centralized baseline, which was trained on all raw data. 
The mesh in Fig.~\ref{fig:large_mesh} was created from the NeRF model learned by agent 1, whose trajectory is marked in red in Fig.~\ref{fig:large_traj}. 
The mesh quality near the living room, directly explored by agent 1, is comparable to the bedroom area, which was reconstructed using information shared by other agents. 
Notably, the multi-agent approach produces fewer floaters, a feature that will be further discussed in relation to sparse input views in the following section. 
These results demonstrate that our multi-agent learning method can scale to larger scenes while maintaining reasonable reconstruction completeness.

\subsection{Training under Different Communication Settings}
\begin{figure}[hbt!]
	\centerline{\includegraphics[trim={5cm 1cm 6cm 1cm},clip,width=19pc]{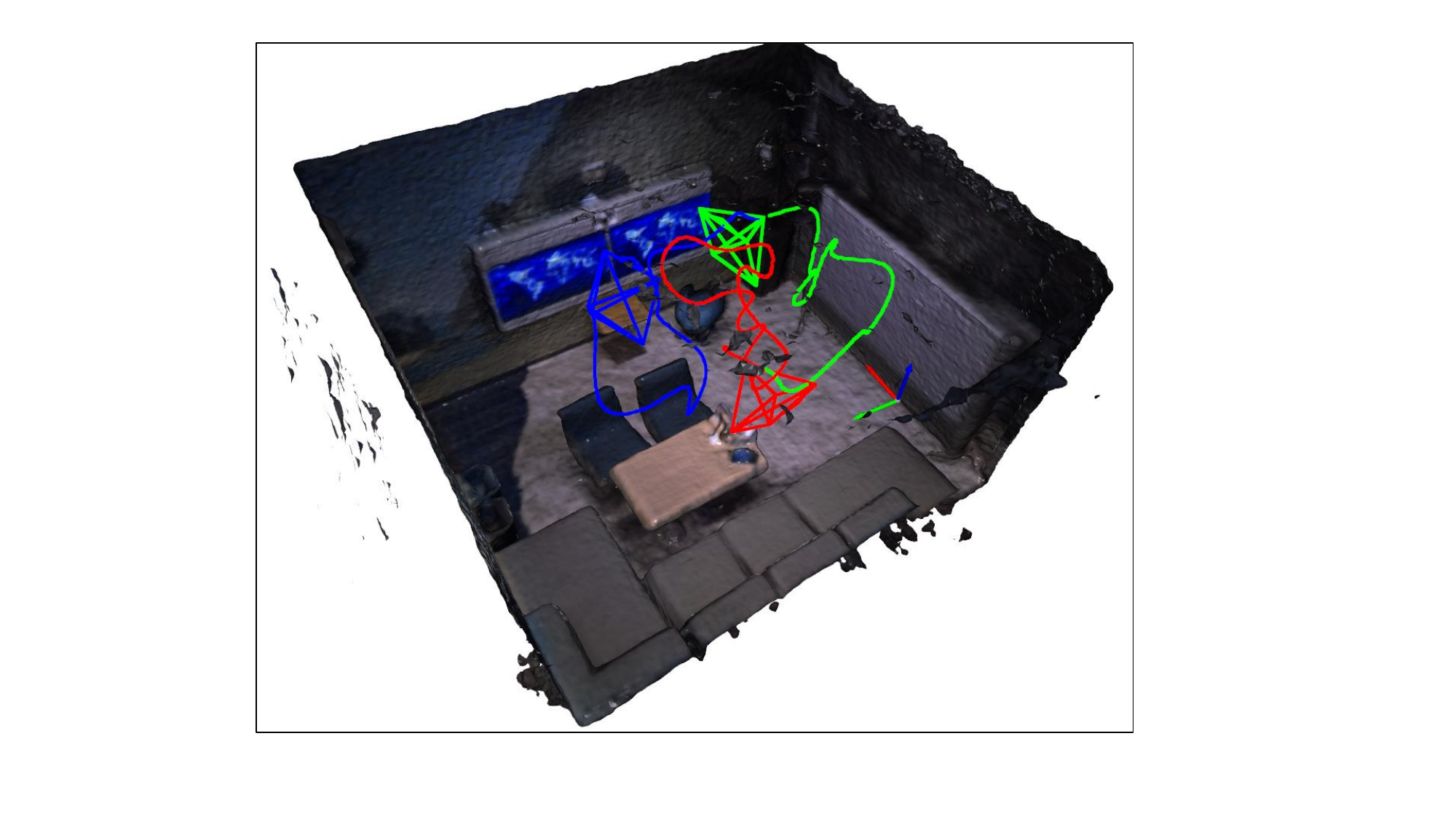}} 
	\caption{Trajectories of the three agents in the experiment, each shown in a different color, simulating a scenario where a group of drones collaboratively explore an indoor environment.}
	\label{fig:traj}
\end{figure}

\begin{figure}[hbt!]
	\centerline{\includegraphics[trim={6cm 1cm 9cm 2cm},clip,width=\columnwidth]{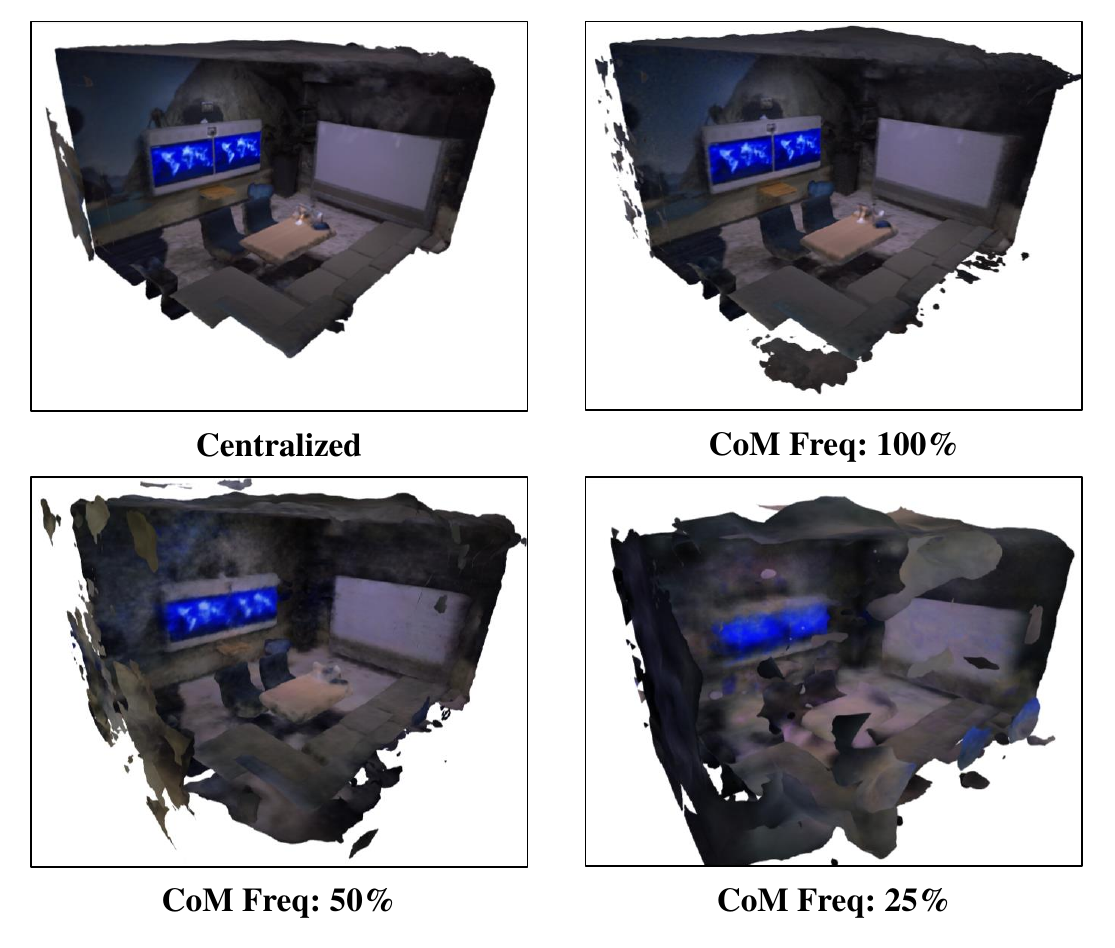}} 
	\caption{Comparison of scene reconstruction quality between multi-agent training under varying communication frequencies and the centralized method. When agents communicate at every iteration (CoM Freq: 100\%), multi-agent training achieves performance comparable to the centralized baseline. As communication frequency decreases, the quality of detailed geometry degrades, but the overall structure of the scene remains intact.}
	\label{fig:com}
\end{figure}

We assess our multi-agent method under varying communication conditions on the Replica dataset. 
At 100\% frequency, agents exchange information with their neighbors at every optimization iteration, representing optimal communication. 
At 25\%, exchanges occur only once every four iterations. 
Three agents were positioned to explore an indoor environment collaboratively, as illustrated in Fig.~\ref{fig:traj}, simulating a multi-drone exploration scenario. 
Fig.~\ref{fig:com} presents the results, using 3D mesh renderings for a thorough evaluation. 
With 100\% communication frequency, multi-agent reconstruction quality closely matches that of the centralized baseline. 
However, as communication frequency decreases, the reconstruction becomes progressively blurrier, likely due to the use of outdated and lower-quality network weights from neighboring agents. 
Despite this, the multi-agent approach still effectively captures the overall structure of the scene.

Communication statistics are summarized in Table \ref{tab:com}.
Each agent received 666 pairs of depth and RGB images, stored as floating-point tensors, during the exploration process. 
An optimization iteration was performed by each agent for every five image pairs received. 
In the centralized training scenario, where all images were transferred to a single agent, the communication cost is substantial. 
In contrast, for distributed training with three agents, each agent only receives model weights from its two neighboring agents per communication iteration. 
This has low bandwidth requirements, making it feasible for real-world deployment. 
Even at a 100\% communication frequency, the total amount of data transferred per agent is far less than in centralized training.

\begin{table}[hbt!]
  \caption{Communication Statistics}
  \centering
  \resizebox{\columnwidth}{!}{
	\begin{tabular}{l r}
		\toprule
		Type &  Size (MB) \\
		\midrule
		Centralized training & 17390.592 \\
		Data received per com iteration per agent& 13.169024 \\
		Total data received (100\% freq) per agent  & 1764.6492 \\
		Total data received (50\% freq) per agent &     882.32461 \\
	Total data received (25\% freq) per agent & 441.16230 \\
		\bottomrule
	\end{tabular}
  }
  \label{tab:com}
\end{table}

\subsection{Sparse Input Views on the Blender and LLFF Datasets}
To explore NeRF learning with sparse input views, we perform both centralized and multi-agent training using the Blender and LLFF datasets. 
The training datasets consist of only four images for Blender scenes and three images for LLFF scenes. 
In centralized NeRF training, a single agent can access all images, while in multi-agent training, images were distributed among multiple agents without overlaps. 
We evaluate the trained models on a test set of 10 images from novel perspectives, including angles from the sides of objects that were not visible to the agents during training.

\begin{figure}[hbt!]
	\centerline{\includegraphics[trim={1cm 2.5cm 1cm 1.5cm},clip,width=\columnwidth]{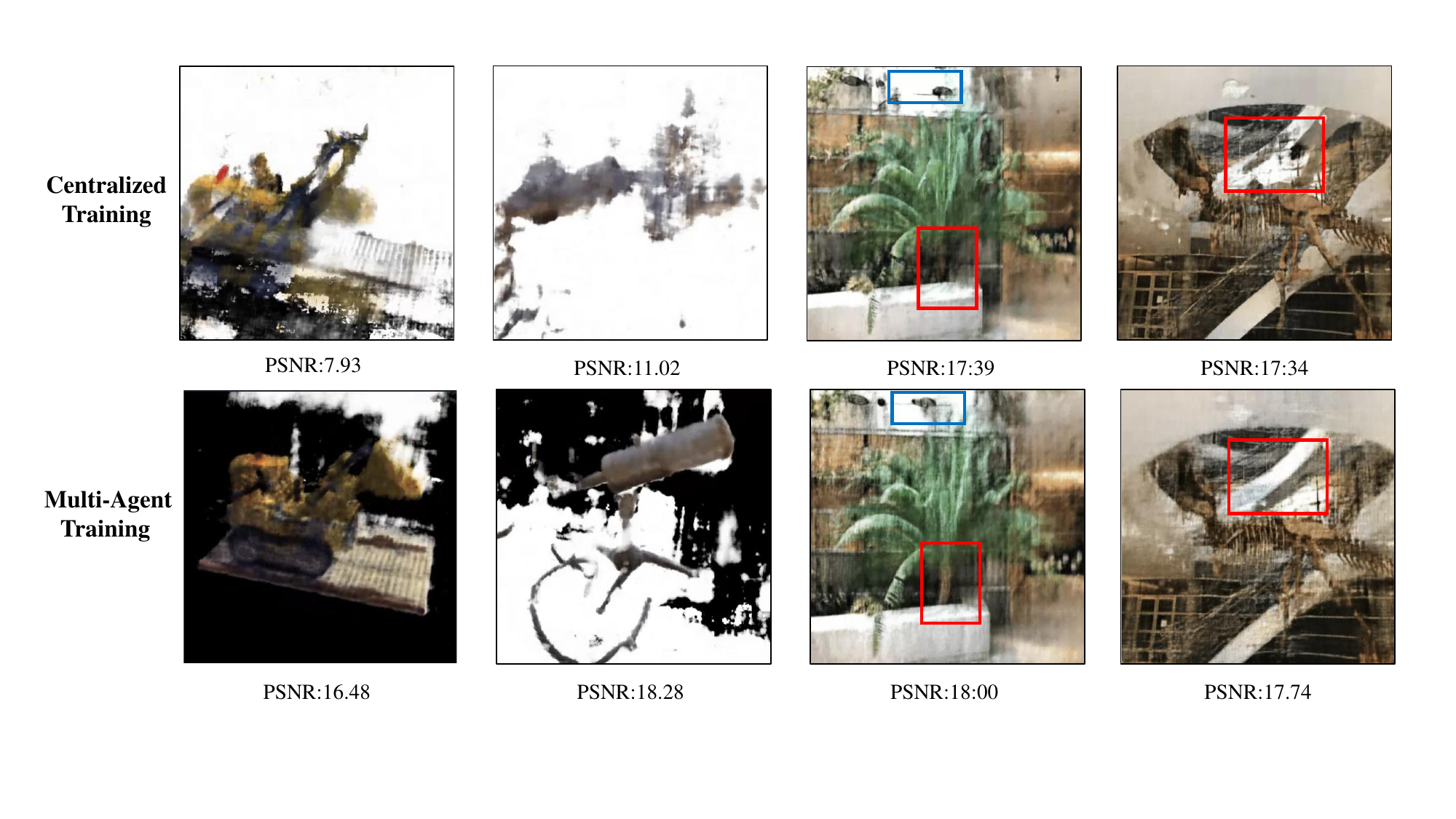}} 
	\caption{With only sparse input views, the centralized NeRF produces inaccurate scene geometries and numerous floaters (highlighted in blue and red rectangles). In contrast, the multi-agent NeRF avoids overfitting and renders significantly higher fidelity novel view perspectives.}
	\label{fig4:blender}
\end{figure}

Table~\ref{tab:BlenderandLLFF} summarizes the quantitative evaluations, with a selection of rendered test images visualized in Fig.~\ref{fig4:blender}. 
When trained on sparse input views, the centralized method demonstrates severe overfitting to the training images, resulting in numerous floaters near the camera (some are highlighted by blue and red rectangles). 
In contrast, multi-agent training effectively recovers correct scene geometries from very few training images, leading to improved visual quality and evaluation scores. 
The capability of multi-agent NeRF learning to mitigate overfitting without additional depth data or pre-trained models presents promising avenues for sparse reconstruction applications.

\begin{table}[hbt!]
    \caption{Blender and LLFF Datasets Results}
    \centering
	\resizebox{\columnwidth}{!}{%
		\begin{tabular}{c ccc ccc} 
			\toprule
			& \multicolumn{3}{c}{LEGO} & \multicolumn{3}{c}{Mic} \\ 
			\cmidrule(lr){2-4} \cmidrule(lr){5-7}
			\# Agents & PSNR $\uparrow$ & SSIM $\uparrow$ & LPIPS $\downarrow$ & PSNR $\uparrow$ & SSIM $\uparrow$ & LPIPS $\downarrow$\\ 
			\midrule
			1 & 7.93 & 0.513 & 0.742 & 11.02 & 0.625 & 0.659 \\ 
			2 & \color{DGreen}\textbf{16.48} & \color{DGreen}\textbf{0.644} & \color{DGreen}\textbf{0.567} & \color{DGreen}\textbf{18.28} & \color{DGreen}\textbf{0.830} & \color{DGreen}\textbf{0.311} \\ 
			\bottomrule
		\end{tabular} 
	}
    \\[1em] 

	\resizebox{\columnwidth}{!}{%
	\begin{tabular}{ccccccc}
		\toprule
		& \multicolumn{3}{c}{Fern} & \multicolumn{3}{c}{T-rex} \\
		\cmidrule(lr){2-4} \cmidrule(lr){5-7}
		\# Agents & PSNR $\uparrow$ & SSIM $\uparrow$ & LPIPS $\downarrow$ & PSNR $\uparrow$ & SSIM $\uparrow$ & LPIPS $\downarrow$ \\
		\midrule
		1 & 17.39 & 0.423 & 0.564 & 17.34 & \color{DGreen}\textbf{0.517} & \color{DGreen}\textbf{0.399} \\
		2 & \color{DGreen}\textbf{18.00} & \color{DGreen}\textbf{0.436} & \color{DGreen}\textbf{0.537} & \color{DGreen}\textbf{17.74} & 0.465 & 0.520 \\
		\bottomrule
	\end{tabular}
  	}
	\label{tab:BlenderandLLFF}
\end{table}

\subsection{Why Does Multi-agent Learning Reduce Overfitting?}
\begin{figure}[hbt!]
	\centerline{\includegraphics[trim={2cm 4cm 11cm 1cm},clip,width=\columnwidth]{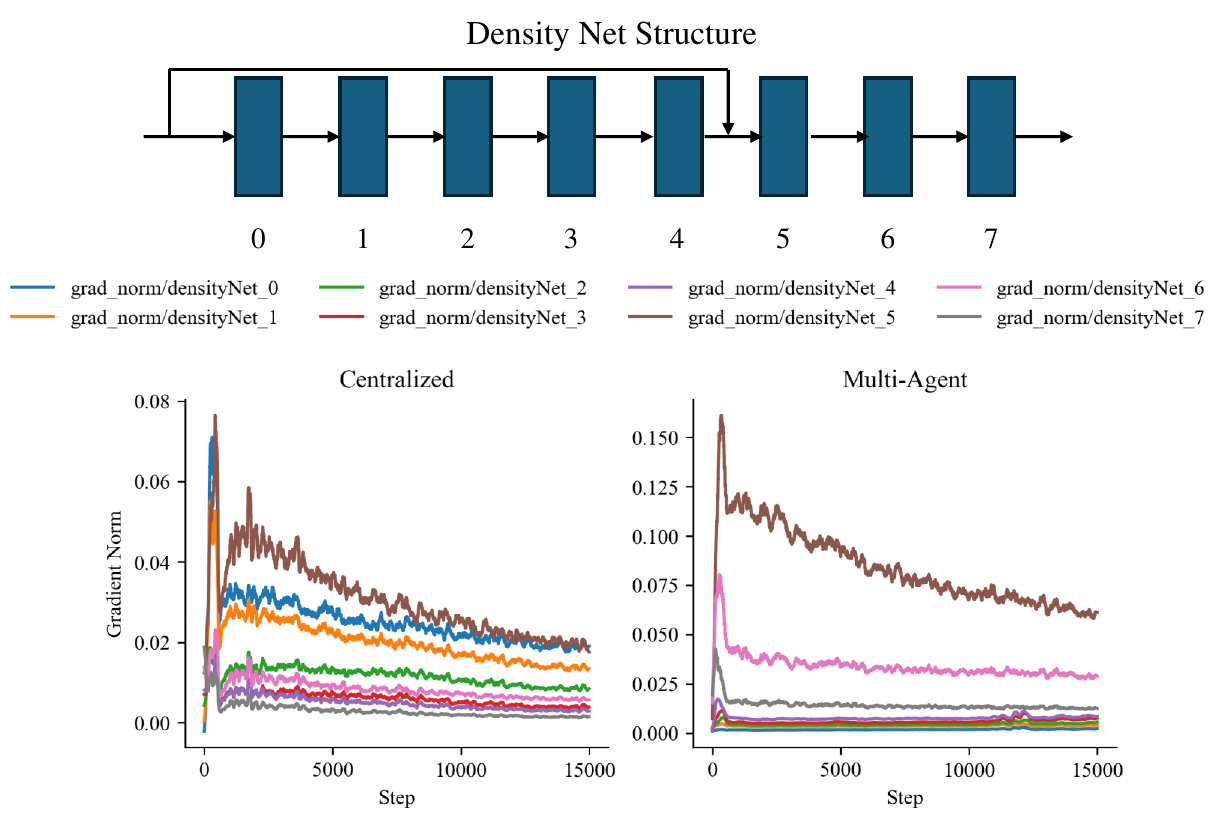}} 
	\caption{The gradient norms for the first eight layers of the MLP networks, for both centralized and multi-agent training of the Blender scene ``Mic''. 
	In multi-agent learning, only layers after the input incorporation receive significant gradients, resulting in a network with fewer parameters. 
	We hypothesize that enforcing consensus among agents drives them to utilize input incorporation to reduce the network capacity, thus mitigating overfitting. 
	}
	\label{fig2:gradient}
\end{figure}

To understand why multi-agent learning reduces overfitting in sparse input training, we draw inspiration from mi-MLP \cite{mi-MLP}, which reduces MLP network capacity by incorporating inputs into each layer, encouraging gradients to flow to only deeper layers (those close to the outputs). 
This facilitates the network to operate with fewer parameters, combating overfitting.
Building on this observation, we hypothesize that the push for consensus among agents promotes a reduction in network capacity through input incorporation. Essentially, the layers preceding input incorporation are bypassed and remain largely unchanged during training 
Intuitively, smaller neural networks reach consensus on their parameters more easily due to the reduced number of parameters.
To investigate this hypothesis, we visualize the gradient norms for the first eight layers of the MLP networks (referred to as ``densityNet'', since they are in charge of learning the volume density) for the Blender scene ``Mic'' in Fig.~\ref{fig2:gradient}.
The gradients across layers in densityNet trained centrally exhibit a similar order of magnitude.
However, multi-agent training only induces notable gradients in the three layers following input incorporation. 
This phenomenon prompts the network to adapt with fewer parameters, mitigating overfitting. 

\section{CONCLUSIONS}\label{Conclusion}
In conclusion, we present a multi-agent NeRF system for collaborative perception, where agents share learned model weights instead of raw data. 
Our distributed learning framework enables more efficient communication, making it suitable for real-world distributed robotic applications. 
Experimental results demonstrate that our approach achieves reconstruction quality comparable to centralized training, while reducing overfitting in sparse input scenarios.

Our method shows great promise, though it currently relies on ideal communication conditions. 
In real-world multi-robot systems, such as drone fleets, communication may face challenges like bandwidth limitations, latency, and interruptions from obstacles or moving agents. 
Despite these potential disruptions impacting reconstruction quality, our experiments demonstrate the method’s strong performance under controlled conditions. 
Future work will focus on addressing these communication challenges to enhance robustness and ensure effective deployment in diverse, real-world scenarios.







\newpage
\bibliographystyle{IEEEtran} 
\bibliography{Mybibfile} 

\end{document}